\documentclass[10pt,twocolumn,letterpaper]{article}

\usepackage[pagenumbers]{cvpr} 

\usepackage{graphicx}
\usepackage{amsmath}
\usepackage{amssymb}
\usepackage{multirow}
\usepackage{booktabs}
\usepackage{makecell}
\usepackage{cuted}
\usepackage{bm}

\makeatletter
\renewcommand\@makefnmark{\hbox{\@textsuperscript{\normalfont\color{red}\@thefnmark}}}
\makeatother

\usepackage[pagebackref,breaklinks,colorlinks]{hyperref}

\usepackage[capitalize]{cleveref}
\crefname{section}{Sec.}{Secs.}
\Crefname{section}{Section}{Sections}
\Crefname{table}{Table}{Tables}
\crefname{table}{Tab.}{Tabs.}

\def\cvprPaperID{****} 
\def\confName{CVPR}
\def\confYear{2023}

\newcommand{\paravspace}{\vspace{-9pt}}

\begin{document}

\title{Progressive Disentangled Representation Learning for Fine-Grained Controllable Talking Head Synthesis}

\author{Duomin Wang \quad Yu Deng \quad Zixin Yin \quad Heung-Yeung Shum \quad Baoyuan Wang \\
	{Xiaobing.AI}\\
}

\vspace{40pt}
\twocolumn[{
\renewcommand\twocolumn[1][]{#1}
\maketitle
\begin{center}
    \captionsetup{type=figure} 
    \includegraphics[width=.9\textwidth]{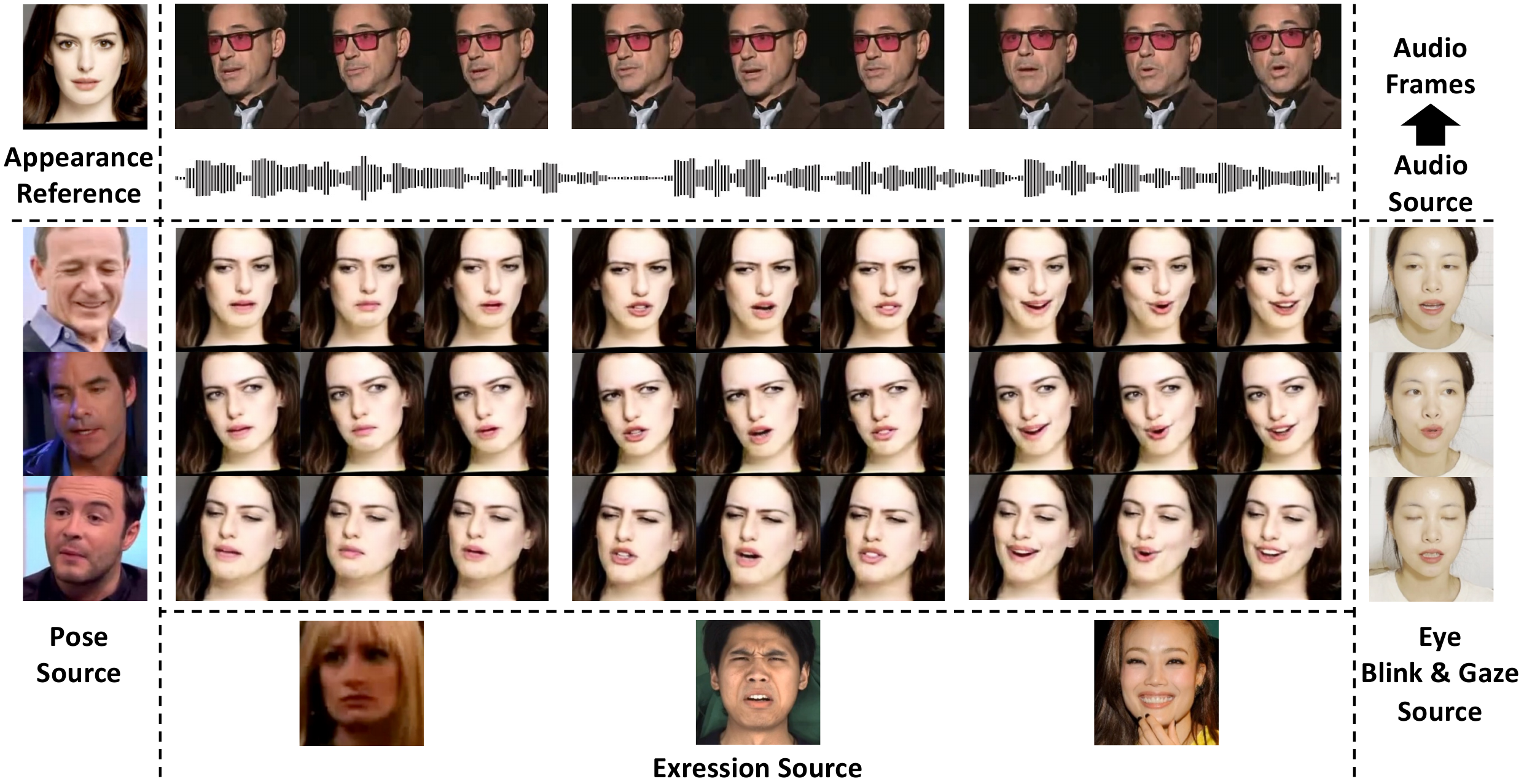}
    \caption{Our method takes an appearance reference as input and generates its talking head with disentangled control over lip motion, head pose, eye gaze\&blink, and emotional expression, where the driving signal of lip motion comes from speech audio, and all other motions are controlled by different videos. As shown, it well disentangles all motion factors and achieves precise control over individual motion.}
    \label{fig:teaser}
\end{center}
}]

\begin{abstract}
We present a novel one-shot talking head synthesis method that achieves disentangled and fine-grained control over lip motion, eye gaze\&blink, head pose, and emotional expression. We represent different motions via disentangled latent representations and leverage an image generator to synthesize talking heads from them. To effectively disentangle each motion factor, we propose a progressive disentangled representation learning strategy by separating the factors in a coarse-to-fine manner, where we first extract unified motion feature from the driving signal, and then isolate each fine-grained motion from the unified feature. We introduce motion-specific contrastive learning and regressing for non-emotional motions, and feature-level decorrelation and self-reconstruction for emotional expression, to fully utilize the inherent properties of each motion factor in unstructured video data to achieve disentanglement. 
Experiments show that our method provides high quality speech\&lip-motion synchronization along with precise and disentangled control over multiple extra facial motions, which can hardly be achieved by previous methods. Project website:\url{https://dorniwang.github.io/PD-FGC/}

\end{abstract}

\section{Introduction}
\label{sec:intro}
Talking head synthesis is an indispensable task for creating realistic video avatars and enables multiple applications such as visual dubbing, interactive live streaming, and online meeting. In recent years, researchers have made great progress in one-shot generation of vivid talking heads by leveraging deep learning techniques. Corresponding methods can be mainly divided into audio-driven talking head synthesis and video-driven face reenactment. Audio-driven methods focus more on accurate lip motion synthesis from audio signals~\cite{suwajanakorn2017synthesizing, chen2018lip, wav2lip, song2022everybody}.
Video-driven approaches~\cite{siarohin2019first,wang2021one} aim to faithfully transfer all facial motions in the source video to target identities and usually treat these motions as a unity without individual control.

We argue that a fine-grained and disentangled control over multiple facial motions is the key to achieving lifelike talking heads, where we can separately control lip motion, head pose, eye motion, and expression, given corresponding respective driving signals. This is not only meaningful from the research aspect which is often known as the disentangled representation learning but also has a great impact on practical applications. Imagining in a real scenario, where we would like to modify the eye gaze of an already synthesized talking head, it could be costly if we cannot solely change it but instead ask an actor to perform a completely new driving motion. Nevertheless, controlling all these factors in a disentangled manner is very challenging. For example, lip motions are highly tangled with emotions by nature, whereas the mouth movement of the same speech can be different under different emotions. There are also insufficient annotated data for large-scale supervised learning to disentangle all these factors. As a result, existing methods either cannot modify certain factors such as eye gaze or expression, or can only change them altogether, or have difficulties providing precise control over individual factors. 

In this paper, we propose {\bf Progressive Disentangled Fine-Grained Controllable Talking Head (PD-FGC)} for one-shot talking head generation with disentangled control over lip motion, head pose, eye gaze\&blink, and emotional expression\footnote{We define the emotional expression as the facial expression that excludes speech-related mouth movement and eye gaze\&blink.}, where the control signal of lip motion comes from audios, and all other motions can be individually driven by different videos. To this end, our intuition is to learn disentangled latent representation for each motion factor, and leverage an image generator to synthesize talking heads taking these latent representations as input. However, it is very challenging to disentangle all these factors given only in-the-wild video data for training. Therefore, we propose to fully utilize the inherent  properties of each motion within the video data with little help of existing prior models. We design a progressive disentangled representation learning strategy to separate each factor control in a coarse-to-fine manner based on their individual properties. It consists of three stages:

1) \textit{Appearance and Motion Disentanglement}.
    We first learn appearance and motion disentanglement via data augmentation and self-driving~\cite{burkov2020neural,zhou2021pose} to obtain a unified motion feature that records all motions of the driving frame meanwhile excludes appearance information. It serves as a strong starting point for further fine-grained disentanglement.

2) \textit{Fine-Grained Motion Disentanglement}. Given the unified motion feature, we learn individual motion representation for lip motion, eye gaze\&blink, and head pose, via a carefully designed motion-specific contrastive learning scheme as well as the guidance of a 3D pose estimator~\cite{danvevcek2022emoca}. Intuitively, speech-only lip motion can be well separated via learning shared information between the unified motion feature and the corresponding audio signal~\cite{zhou2021pose}; eye motions can be disentangled by region-level contrastive learning that focuses on eye region only, and head pose can be well defined by 3D rigid transformation.    

3) \textit{Expression Disentanglement}. Finally, we turn to the challenging expression separation as the emotional expression is often highly tangled with other motions such as mouth movement. We achieve expression disentanglement via decorrelating it with other motion factors on a feature level, which we find works incredibly well. An image generator is simultaneously learned for self-reconstruction of the driving signals to learn the semantically-meaningful expression representation in a complementary manner. 

In summary, our contributions are as follows: \textbf{1)} We propose a novel one-shot and fine-grained controllable talking head synthesis method that disentangles appearance, lip motion, head pose, eye blink\&gaze, and emotional expression, by leveraging a carefully designed progressive disentangled representation learning strategy. \textbf{2)} Motion-specific contrastive learning and feature-level decorrelation are introduced to achieve desired factor disentanglement. \textbf{3)} Trained on unstructured video data with limited guidance from prior models, our method can precisely control diverse facial motions given different driving signals, which can hardly be achieved by previous methods.


\section{Related work}
\label{sec:related}
\paragraph{Audio-driven talking head synthesis.} 
Audio-driven talking head synthesis~\cite{bregler1997video, brand1999voice} aims to generate portrait images with synchronized lip motions to the given speech audios. The majority of works~\cite{lu2021live,zhu2018arbitrary,suwajanakorn2017synthesizing, chen2018lip, wav2lip, song2022everybody} focus on controlling only the mouth region and leave other parts unchanged. Some recent works enable control over more facial properties such as eye blink and head pose~\cite{yi2020audio, thies2020nvp, chen2020talking, zhou2020makelttalk, zhang2020apb2facev2, zhang2021facial}.
More recently, several methods~\cite{vougioukas2020realistic, evp2021, ji2022eamm, Liang_2022_CVPR} try to introduce emotional expression variations into the synthesis process as it is a crucial property for vivid talking head generation. However, integrating expression control into talking-head synthesis is very challenging due to the lack of expressive data. Some methods~\cite{wang2020mead,evp2021,ji2022eamm} build on manually collected emotional talking head dataset~\cite{wang2020mead}, yet they cannot well generalize to large-scale scenarios due to the limited data coverage. A recent work GC-AVT~\cite{Liang_2022_CVPR} leverages in-the-wild data for expressive talking head synthesis. They achieve disentangled control over expression by introducing mouth-region data augmentation to separate lip motion and other facial expressions. Different from them, we leverage a feature-level decorrelation to disentangle the two factors. Moreover, our method can synthesize arbitrary talking head with a disentangled control over lip motion, head pose, eye gaze\&blink, and expressions, while previous methods cannot achieve individual control over all these factors.

\paravspace
\paragraph{Video-driven face reenactment.}
Video-driven face reenactment targets faithful facial motion transfer between a driving video and a target image. The literature can be mainly divided into warping-based methods~\cite{siarohin2019first, gu2020flnet, zeng2020realistic, x2face2018, yao2021one, wang2021one, wang2022latent, ren2021pirenderer, hong2022depth, drobyshev2022megaportraits} and synthesis-based approaches~\cite{thies2016face2face, wu2018reenactgan, kim2018deep, pumarola2018ganimation, burkov2020neural, xiang2020one, sun2022landmarkgan, liu2021li}. The warping-based methods predict warping flows between the source and target frames to transform target images or their extracted features to align with source motions. The synthesis-based methods instead learn intermediate representations from input images and directly send them to a generator for image synthesis. The representations can be landmarks~\cite{xiang2020one, sun2022landmarkgan, liu2021li}, 3D face model parameters or meshes~\cite{kim2018deep, ren2021pirenderer, thies2016face2face, cao2021unifacegan}, or latent features extracted from images~\cite{burkov2020neural, wang2022latent, bounareli2022finding}. Some recent methods~\cite{lin20223d, yin2022styleheat} also exploit prior knowledge from a pre-trained 2D GAN~\cite{karras2020analyzing} for animating face images. Our proposed method also builds on a synthesis-based approach, where we learn disentangled latent representations for multiple facial motions by our designed progressive disentangled representation learning strategy. In addition, different from video-driven approaches, we control the lip motion via audio signals.

\paravspace
\paragraph{Disentangled representation learning on the face.}
Disentangled representation learning for faces is a longstanding task and has been widely explored in the literature. Plenty of works~\cite{chen2016infogan, lin2019infogan, Wei_2021_ICCV, nguyen2019hologan,higgins2016beta, kim2018disentangling, chen2018isolating, ding2020guided} focus on unsupervised representation learning, where InfoGAN~\cite{chen2016infogan} and $\beta$-VAE~\cite{higgins2016beta} are two representative works. However, these unsupervised methods cannot guarantee meaningful latent representations well aligned with human perceptions~\cite{locatello2019challenging}. More recently, several methods~\cite{shen2020interfacegan, shen2020interpreting, zhu2020improved, wu2021stylespace, zheng2021unsupervised, shen2021closed} explore latent space editing of a pre-trained 2D GAN~\cite{karras2020analyzing} with the help of certain classifiers to achieve disentangled control over desired facial properties. Nevertheless, their controllability is often confined by the linear classifiers and the data distribution of the pre-trained generator. Some methods~\cite{donahue2017semantically, deng2020disentangled, ghosh2020gif, ren2021pirenderer, pumarola2018ganimation, xiao2018elegant} leverage more powerful prior knowledge such as 3D face model~\cite{paysan20093d} or expression model~\cite{pumarola2018ganimation} to guide the representation learning, and develop specific training schemes~\cite{donahue2017semantically, xiao2018elegant} on structured data to achieve desired factor disentanglement. We achieve disentangled representation learning via a carefully designed progressive training scheme on videos and introduce certain prior models~\cite{danvevcek2022emoca} to help with accurate factor control.


\begin{figure*}[t]
  \centering
  \includegraphics[width=1\linewidth]{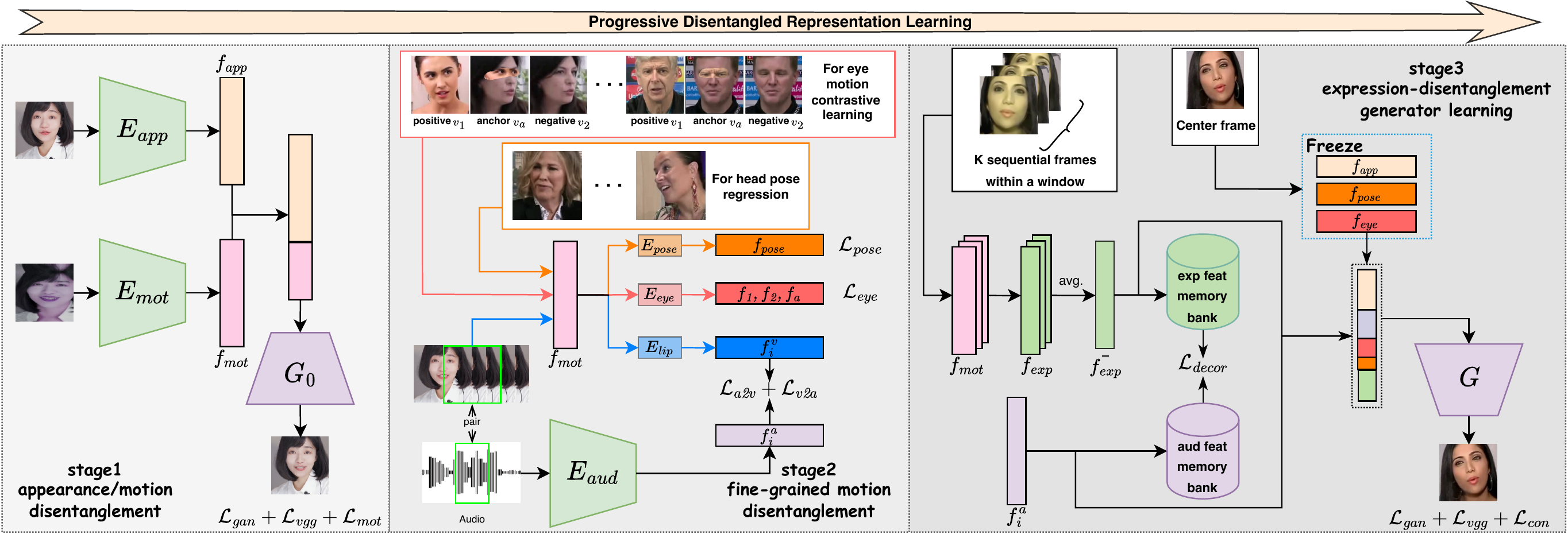}
    
  \caption{The overview of our method. We achieve factor disentanglement for different facial motions via a progressive disentangled representation learning strategy. We first disentangle appearance with all facial motions to obtain a unified motion feature for further fine-grained disentanglement. Then, we separate each fine-grained motion feature from the unified motion feature via motion-specific contrastive learning and the help of a 3D prior model. Finally, we disentangle expression with other motions by feature-level decorrelation and simultaneously learn an image generator for controllable talking head synthesis.}
  \label{fig:pipeline}
\end{figure*}

\section{Method}
\label{sec:method}
Given an image of an arbitrary person, our goal is to synthesize a talking-head video of it, where we can separately control different facial motions in each frame, including lip motion, head pose, eye gaze\&blink, and emotional expression. We expect the lip motion to be derived from an audio clip and other motions from different respective driving videos.
To this end, we propose to represent all controllable facial motions along with their appearance by disentangling latent representations of the input visual and audio signals and learning a corresponding image generator to synthesize desired talking heads from them. We introduce a progressive disentangled representation learning scheme to learn the latent representations in a coarse-to-fine manner, as shown in~\cref{fig:pipeline}. We first disentangle the appearance with the facial motions to obtain a unified motion representation that records all motion information (\cref{subsec:appearce_motion}). Then, we isolate each fine-grained facial motion, except the expression, from the unified motion feature via motion-specific contrastive learning (\cref{subsec:motion_disentagle}). Finally, we separate the expression from other motions via feature-level decorrelation, and simultaneously learn the image generator for fine-grained controllable talking head synthesis (\cref{subsec:exp_ctrl_gan}).

\subsection{Appearance and Motion Disentanglement}
\label{subsec:appearce_motion}
We argue that to achieve disentanglement over multiple fine-grained motion factors, a primary thing to do is to learn a unified motion representation that records all kinds of motion information meanwhile excludes appearance (\ie identity) information. Such a unified motion feature serves as a strong starting point for further fine-grained factor disentanglement  from it. To this end, we follow~\cite{burkov2020neural} to disentangle appearance and facial motions. 

Specifically, an appearance encoder $E_{app}$ and a motion encoder $E_{mot}$ are introduced to extract corresponding features from an appearance image and a driving frame, respectively. An extra generator $G_0$ is applied to synthesize a face image with the identity of the appearance image and the facial motion of the driving frame. Self-driving and reconstruction are applied to learn the whole pipeline, where data augmentation is introduced to the motion branch to force the motion encoder to neglect appearance variations and only focus on motion extraction. To further improve the accuracy of extracted  motion feature, we introduce a motion reconstruction loss on top of the training losses in~\cite{burkov2020neural}:
\begin{equation}
  \mathcal{L}_{mot} = \|\bm{\mathcal{\phi}}(I_0)- \bm{\mathcal{\phi}}(I_g)\|_2 + \|\bm{\mathcal{\psi}}(I_0)- \bm{\mathcal{\psi}}(I_g)\|_2,
  \label{eq:loss_mot}
\end{equation}
where $\bm{\mathcal{\phi}}(\cdot)$ and $\bm{\mathcal{\psi}}(\cdot)$ are features extracted by the 3D face reconstruction network and the emotion network of~\cite{danvevcek2022emoca}, $I_0$ is the image synthesized by $G_{0}$ given appearance and motion features, and $I_g$ is the ground truth image. The above training scheme helps us to learn a unified motion feature that faithfully represents all facial movements, which further helps us to achieve fine-grained motion disentanglement to be described in the following sections.

\subsection{Fine-Grained Motion Disentanglement}
\label{subsec:motion_disentagle}
Based on the unified motion feature from the previous stage, we introduce three extra encoders to further extract fine-grained motion features from it, including lip motion feature, eye gaze\&blink feature, and head pose feature. The key intuition is to design motion-specific contrastive learning based on the unique property of each individual motion or to leverage the guidance of a prior model if a motion can be well described by it. We do not separate expression in this stage as it can be highly tangled with other factors. We leave its disentanglement in the final stage (\cref{subsec:exp_ctrl_gan}) where all other factors are effectively separated.

\paravspace
\paragraph{Lip motion contrastive learning.} Lip motions can be well separated from other motions by exploring shared information between the unified motion feature and the corresponding speech audio, as shown by previous method~\cite{zhou2021pose}. Therefore, we follow~\cite{zhou2021pose} to learn the lip motion feature with audio-visual contrastive learning.
Given a set of video frames $\{v_i\}$ and their corresponding audio signals $\{a_i\}$, we introduce a lip motion encoder $E_{lip}$ and an audio encoder $E_{aud}$, and extract lip motion features $\{f^v_i\} = \{E_{lip}\circ E_{mot}(v_i)\}$ and audio features $\{f^a_i\} = \{E_{aud}(a_i)\}$ via the two networks, where $E_{mot}$ is the pre-trained motion encoder from the previous stage. We then construct a positive audio-video pair $(f^a_i,f^v_i)$ and $K$ negative audio-video pairs $(f^a_i,f^v_k), k \neq i$ for each sampled audio feature $f^a_i$, and vice versa. We enforce the InfoNCE loss~\cite{oord2018representation} following~\cite{zhou2021pose} to maximize the similarity between positive pairs and minimize the similarity between negative pairs:
\begin{equation}
\resizebox{0.9\hsize}{!}{$\mathcal{L}_{a2v} = -{\rm log}[\frac{\exp(\mathcal{S}(f^a_i, f^v_i))}{\exp(S(f^a_i, f^v_i)) + \sum_{k=1}^K \exp(\mathcal{S}(f^a_i, f^v_k))}],
  \label{eq:cl_a2v}$}
\end{equation}
\begin{equation}
\resizebox{0.9\hsize}{!}{$ \mathcal{L}_{v2a} = -{\rm log}[\frac{\exp(\mathcal{S}(f^v_i, f^a_i))}{\exp(\mathcal{S}(f^v_i, f^a_i)) + \sum_{k=1}^K \exp(\mathcal{S}(f^v_i, f^a_k))}],$}
  \label{eq:cl_v2a}
\end{equation}
where $\mathcal{S}(\cdot,\cdot)$ is the cosine similarity. This loss ensures that lip motion features predicted by $E_{lip}$ and $E_{aud}$ are close to each other for corresponding video frames and audio. Since the audio signal merely contains lip motion information, it helps with better factor disentanglement. Moreover, we can leverage the audio encoder for audio-driven lip motion synthesis for our controllable talking head. 

\paravspace
\paragraph{Eye motion contrastive learning.} 
Eye motions, including eye gaze and blink, are local movements and have limited influence on other facial regions. Therefore, if we substitute the eye region of a person with that of another person to composite a new image, the extracted eye motion feature from it should be identical to that of the latter person. 
Based on this observation, we design a dedicated contrastive learning scheme to disentangle the eye motions.

Specifically, given two driving frames, namely $v_1$ and $v_2$, we construct an anchor frame $v_a$ by compositing the eye region\footnote{We use an off-the-shelf method~\cite{yu2021heatmap} to detect eye landmarks, and warp the eye region of $v_1$ to align with that of $v_2$.} of $v_1$ and other regions of $v_2$, as shown in \cref{fig:pipeline}. We introduce an extra encoder $E_{eye}$ to extract eye motion features $f_1$, $f_2$, and $f_a$ from the corresponding unified motion feature of the above frames, and construct a positive pair $(f_1,f_a)$ and a negative pair $(f_2,f_a)$. We then enforce a similar InfoNCE loss between these pairs to learn $E_{eye}$:
\begin{equation}
  \mathcal{L}_{eye} = -{\rm log}[\frac{\exp(\mathcal{S}(f_1, f_a))}{\exp(\mathcal{S}(f_1, f_a)) + \exp(\mathcal{S}(f_2, f_a))}].
  \label{eq:tcl_eye}
\end{equation}
It helps the eye motion encoder to only focus on eye region motions and neglect the variance of other regions.

\paravspace
\paragraph{Head pose learning.} 
Since head pose can be well defined by a 6D parameter consisting of three Euler angles (\ie pitch, yaw, roll) and 3D translations, we propose to directly regress them via a head pose encoder $E_{pose}$ with the guidance of a 3D face prior model:
\begin{equation}
  \mathcal{L}_{pose} = |\mathcal{P}_{pred} - \mathcal{P}_{gt}|_1,
  \label{eq:loss_pose}
\end{equation}
where $ \mathcal{P}_{pred}$ is the predicted pose parameter by $E_{pose}$, and $\mathcal{P}_{gt} $ is the ground truth pose parameter obtained by the off-the-shelf 3D face reconstruction model~\cite{danvevcek2022emoca}.

\subsection{Expression Disentanglement}
\label{subsec:exp_ctrl_gan}
The challenge for expression disentanglement is two-fold. On one hand, the emotional expression can be highly tangled with other motions (\eg mouth movements can be different under different emotions even if the speech contents are identical), which makes it difficult to design motion-specific contrastive learning as done previously. On the other hand, existing expression estimators~\cite{mollahosseini2017affectnet,deng2019accurate,danvevcek2022emoca} usually include other motion information in their expression representation, which cannot provide accurate guidance for the expression disentanglement in our scenario. To tackle these challenges, we propose a feature-level decorrelation strategy to disentangle the expression with other motions, along with a self-reconstruction of the driving frame to learn precise expression representation in a complementary manner. The hypothesis behind this is that if an extracted expression feature is independent of the features of other motions, meanwhile its combination with the others can still faithfully reconstruct all facial motions in the driving signal, then it is a precise latent representation of the ground truth expression. We describe the learning strategies in detail below.

\paravspace
\paragraph{In-window decorrelation.}
We observe that the expression variation in a video sequence is usually less frequent than the changes in other motions. Therefore, if we take the average expression feature within a time window, the other motion information stored in certain dimension of the expression feature should be averaged out, leading to a clean expression feature uncorrelated with other motions. Therefore, given a driving frame, we define a window of size $K$ around it and augment the frames within the window with random rotation, scaling, and color jittering. We then extract the expression features 
from their corresponding unified motion features via an expression encoder $E_{exp}$, and calculate their average feature as the expression feature for the center driving frame. The average feature will be then sent into a generator $G$ for image synthesis and self-reconstruction described in the following paragraph\footnote{During inference, we use the expression feature of each frame instead of the average one, as we find the expression encoder learned following our training strategy can well disentangle expression with other motions.}.

\paravspace
\paragraph{Lip-motion decorrelation.}
We further introduce a lip motion decorrelation loss to achieve better expression disentanglement by forcing independence between the expression feature and the lip motion feature (\ie audio feature): 
\begin{equation}
  \mathcal{L}_{decor} = \frac{1}{D}\sum_{B, D}cor(\bar{F^e}, F^a)^2,
  \label{eq:loss_decor}
\end{equation}
where $\bar{F^e} \in \mathbb{R}^{B\times D}$ is a matrix consisting of average expression features within a batch of size $B$, $F^a \in \mathbb{R}^{B\times D}$ is the corresponding audio feature matrix, $D$ is the feature dimension, and $cor(\cdot,\cdot)$ calculates the feature dimension correlation between the two matrices. In practice, computing the correlation between two variables requires a large batchsize to reach enough accuracy. However, it is difficult to maintain such a large batchsize during training due to memory limitation. To tackle this problem, we maintain two memory banks for the expression feature and the audio feature to compute the correlation, instead of using only the current batch of features. The memory bank always keeps $M$ latest features 
inside to compute \cref{eq:loss_decor} during training, where $M $ is much larger than the batchsize of each iteration. The gradient will only back-propagate through the current batch of features to update the network weights.

\paravspace
\paragraph{Complementary learning via self-reconstruction.} The above two decorrelation strategies ensure feature independence between expression and other motions, yet the extracted expression feature still lacks semantic meaning. Therefore, we leverage an image generator $G$ to take the expression feature along with the features of appearance and other motions as input, and synthesize an image with desired facial motions via self-reconstruction of the driving frame, as shown in \cref{fig:pipeline}. In order to faithfully reconstruct the driving frame, the expression encoder is forced to learn complementary information that is not included in all other motion features, which is exactly the expression information. We enforce multiple losses to learn the expression encoder $E_{exp}$ and the image generator $G$:
\begin{equation}
  \mathcal{L}_{vgg} = \sum_{i=1}^N \lVert VGG_i(I_f) - VGG_i(I_g) \rVert_1,
  \label{eq:vgg}
\end{equation}
where $VGG_i(\cdot)$ is the feature map of the $i's$ layer in a pre-trained VGG19~\cite{simonyan2014very} network. We also adopt the adversarial loss and the discriminator feature matching loss following~\cite{burkov2020neural} to improve the synthesized image quality.

In addition, to ensure that the synthesized image well follows all facial motions of the driving frame, we further introduce a motion-level consistency loss:

\begin{equation}
\begin{split}
  \mathcal{L}_{con} &= \exp(-\mathcal{S}(\bm{\mathcal{V}}_{lip}(I_f), E_{aud}(a_g)))\\
  &   + \lVert\bm{\mathcal{G}}(I_f), \bm{\mathcal{G}}(I_g)\rVert_1 + \mathcal{L}_{mot},
  \label{eq:loss_consist}
\end{split}
\end{equation}
where $E_{aud}$ is our audio encoder learned in the previous stage, $ \bm{\mathcal{V}}_{lip} $ is a pre-trained encoder to extract lip motion features from images, $\mathcal{G}(I)$ is a gaze estimator~\cite{abdelrahman2022l2cs}, and $\mathcal{S}(\cdot,\cdot)$ is the cosine similarity; $I_f$, $I_g$ and $a_g$ are our synthesized image, ground truth image, and audio, respectively; $\mathcal{L}_{mot}$ is the motion reconstruction loss defined in~\cref{eq:loss_mot}.

The above self-reconstruction process, together with the feature-level decorrelation strategy, helps to disentangle the expression feature from the unified motion feature. Moreover, the image generator $G$ learned in this step naturally achieves disentangled and controllable talking head synthesis with all disentangled motion features and the appearance feature as input. We, therefore, take $G$ as our final image generator for talking head synthesis.

\section{Experiments}
\label{sec:Experiments}

\begin{table*}\small
\centering
\caption{Quantitative comparison for audio-driven talking head synthesis on VoxCeleb2~\cite{Chung18b} 
and Mead~\cite{wang2020mead}. \dag: The LSE-C value for the training data of each method is shown in the bracket as a reference.}
\begin{tabular}{@{}l|ccccc|ccccc@{}}
\toprule
\multicolumn{1}{c}{\multirow{2}{*}{Method}} & \multicolumn{5}{c}{VoxCeleb2}                & \multicolumn{5}{c}{Mead}                \\
\cline{2-11}
\multicolumn{1}{c}{}     & \!FID$\downarrow$\!  & \!LSE-C$\uparrow$\! & \!NLSE-C$^\dag\downarrow$\! & \!LMD$_m\downarrow$\! & \!LMD$\downarrow$\! & \!FID$\downarrow$\!  & \!LSE-C$\uparrow$\! & \!NLSE-C$^\dag\downarrow$\! & \!LMD$_m\downarrow$\! & \!LMD$\downarrow$\! \\
\hline
GT     &  -    &   7.35    &   0&  0 & 0    &   -    &     1.76    &      0     &  0 &    0    \\
\midrule
Wav2Lip~\cite{wav2lip}  &   -  &  \bf{9.23} & 0.183(7.80) &  2.54   &  -  &  -   & \bf{9.17} & 0.176(7.80) & 2.54   &  -    \\
MakeItTalk~\cite{zhou2020makelttalk}    &   19.47  &    2.03 & 0.724(7.35) &   2.82    &   6.39    &   68.35    & 3.50 & 0.524(7.35) & 2.68 &  2.56 \\
PC-AVS~\cite{zhou2021pose}       &   14.36    &     8.21 &  0.117(7.35)  &   1.59    &   2.52  &  62.85    &  8.04 & 0.094(7.35) &  2.28 & 1.92    \\
EAMM Neutral~\cite{ji2022eamm}     &  26.06   &   4.75 & 1.699(1.76)  &  2.29   & 4.18 &   50.36  &    5.34    &   2.03(1.76)   &   2.32  &    2.25      \\
EAMM Emo~\cite{ji2022eamm}   &  27.20     &   4.55 & 1.585(1.76)  &2.29     &    4.29    &   \bf{50.49}   &   5.23 & 1.972(1.76) &  2.28  &  2.27 \\
\midrule
PD-FGC (Ours)     &  \bf{12.99}    &  7.26 & \bf{0.012}(7.35) & \bf{1.15}   &   \bf{1.93}  &  73.80   &   7.24  & \bf{0.015}(7.35)  & \bf{1.65}  &  \bf{1.84}  \\
\bottomrule
\end{tabular}
\label{tab:table_lip_motion}
\end{table*}

\paragraph{Implementation details.}
We train our model on VoxCeleb2~\cite{Chung18b} dataset and evaluate it on both VoxCeleb2 and Mead~\cite{wang2020mead} dataset. All video frames are aligned following the official annotations~\cite{Chung18b} and resized to $224\times224$. Corresponding audios are extracted from the original videos and converted to Mel-sectrograms. Our appearance encoder $E_{app}$ is implemented as a ResNet50~\cite{he2016deep}, and the motion encoder $E_{mot}$ takes the same structure as~\cite{bulat2017far}. The audio encoder $E_{aud}$ adopts a ResNetSE34~\cite{jung2022pushing} structure. The encoder for each fine-grained motion factor, including lip motion, head pose, eye gaze\&blink, and expression, is implemented as an MLP with ReLU activations. The image generator $G_0$ in the first stage and the final image generator $G$ are both based on StyleGAN2~\cite{karras2020analyzing}. The dimensions of the appearance feature and the unified motion feature are set to  $2,048$ and $512$, respectively. The dimensions of each fine-grained motion features are $500$, $6$, $6$, and $30$ for lip motion (audio), head pose, eye gaze\&blink, and expression, respectively.  We implement our framework using PyTorch~\cite{paszke2019pytorch}, and train it on $8$ Tesla V100 GPUs with $32$GB memory, using a batchsize of 16 for 50 epochs. See the supplementary materials for more details.

\paravspace
\paragraph{Baselines.} We compare our method with existing talking head synthesis methods: {\bf Wav2Lip}~\cite{wav2lip} that allows only mouth region control; {\bf MakeItTalk}~\cite{zhou2020makelttalk} that further introduces random eye blinks and audio-aware head poses; {\bf PC-AVS} \cite{zhou2021pose} with controllable head pose and lip motion; and {\bf EAMM} \cite{ji2022eamm} with disentangled control over lip motion, head pose, and expression. A recent GC-AVT~\cite{Liang_2022_CVPR} also achieves expressive talking head synthesis. We do not compare with it since its code is unavailable yet.

\begin{table}[]\small
\centering
\caption{Comparison for expression and pose control accuracy.}
\begin{tabular}{@{}l|cc|c@{}}
\toprule
\multicolumn{1}{c}{\multirow{2}{*}{Method}} & \multicolumn{2}{c}{Expression$\downarrow$}  & \multicolumn{1}{c}{Pose$\downarrow$}      \\
\multicolumn{1}{c}{} & VoxCeleb2 & Mead & VoxCeleb2 \\
\hline
  
  PC-AVS~\cite{zhou2021pose} &   0.202  &      0.245     & 0.0038  \\
  EAMM emo~\cite{ji2022eamm} &   0.196  &     0.245   &   0.0196    \\
  EAMM neutral~\cite{ji2022eamm} &  0.192   &    0.248    & 0.0203   \\
  \midrule
  PD-FGC (Ours) &  \bf{0.156}  &      \bf{0.188}   &  \bf{0.0016}   \\
      
\bottomrule

\end{tabular}
\label{table_crx_exp_pose}
\end{table}

\subsection{Quantitative Evaluation} \label{sec:quant}
We evaluate the image generation quality as well as factor control accuracy of different methods on a self-driving setting, where we use the first frame in a test video to provide appearance and use the following audio and video frames to provide lip motion and other motions, respectively. We use the Fr\'echet Inception Distances ({\bf FID})~\cite{heusel2017gans} between the synthesized images and the ground truth driving frames to evaluate the image quality. For the accuracy of motion control, we leverage several metrics. We first calculate the facial landmark distance ({\bf LMD})~\cite{chen2019hierarchical} between the synthesized images and the ground truth to evaluate the overall motion control accuracy. We further calculate the mouth region landmark distance ({\bf LMD}$_m$) to evaluate the accuracy of lip motion control. We also adopt the Lip Sync Error Confidence ({\bf LSE-C}, also known as Sync$_{conf} $)~\cite{wav2lip} to evaluate the lip motion synchronization with the driving audio.
Nevertheless, the {\bf LSE-C} value of a method is strongly correlated with its training data, which makes it unfair when comparing methods trained on different data. A recent method~\cite{yao2022dfa} also indicates that the {\bf LSE-C} difference between synthesized images of a method and its training data, rather than the absolute value, better reveals lip motion synchronization. Therefore, we propose a normalized confidence score {\bf NLSE-C} for a fair comparison:
\begin{equation}
 {\bf NLSE\mbox{-}C} = \frac{ {\bf LSE\mbox{-}C}^{gen} -  {\bf LSE\mbox{-}C}^{gt}}{{\bf LSE\mbox{-}C}^{gt}},
  \label{eq:nlsec}
\end{equation}
where {\bf LSE-C}$^{gen}$ is the {\bf LSE-C} score of generated images, and {\bf LSE-C}$^{gt}$ is the corresponding score of the training data. The new {\bf NLSE-C} measures the relative difference between the generated images and their training data, which better reveals if the synthesized lip motions are reasonable (\ie close to the training data distribution) or not. We find that this new metric better aligns with human perception.

The quantitative results are shown in \cref{tab:table_lip_motion}. Our method yields the best motion control accuracy in terms of  {\bf NLSE-C}, {\bf LMD}, and {\bf LMD}$ _m $. We also show competitive image generation quality with other methods. Since Wav2Lip only generates mouth region and copies other regions from input images, we do not evaluate its FID and LMD which can be unfair to other methods.

We further compare our method with the others on expression and head pose control accuracy. For the head pose evaluation, we follow the same self-driving setting as described above, and use a 3D face reconstructor~\cite{deng2019accurate} to extract head pose parameters from the synthesized images and calculate their difference ({\bf MSE}) with the ground truth. For the expression evaluation, we find the self-driving setting cannot well evaluate expression controllability as the appearance reference usually contains similar expressions with the driving frames if they come from the same video clip. Therefore, we conduct a cross-video setting where we use appearance reference and driving frames from different video clips for image synthesis (the driving audio is still from the video clip of the appearance reference). We use a 3D face reconstructor~\cite{deng2019accurate} to extract expression parameters in the synthesized images, and compare their difference ({\bf MSE}) with those of the driving frames to evaluate the control accuracy. As shown in ~\cref{table_crx_exp_pose}, we achieve the lowest expression and pose control error largely outperforming previous methods. More details are in the supplementary materials.

\begin{table}[]\small
\centering
\caption{User study on talking head synthesis.}
\setlength{\tabcolsep}{1.5mm}{
\begin{tabular}{@{}l|ccc@{}}
\toprule
Method & \makecell[c]{Lip Sync \\Quality$\uparrow$}  & \makecell[c]{Expression \\Quality$\uparrow$}  & \makecell[c]{Facial Motion \\ Driving Naturalness$\uparrow$} \\
\hline
Wav2Lip~\cite{wav2lip} & 3.50 & 1.36 & 1.88 \\
MakeItTalk~\cite{zhou2020makelttalk} & 1.81 & 1.89 & 2.65 \\
PC-AVS~\cite{zhou2021pose}  & {\bf4.46} & 3.04 & 3.72\\
Eamm~\cite{ji2022eamm}  & 1.92 & 1.77 & 1.56 \\
\midrule
PD-FGC (Ours)  & 4.44 & {\bf4.38} & {\bf4.27}\\
\bottomrule
\end{tabular}
}
\label{table_user_study}
\end{table}

\subsection{Qualitative Evaluation}

\paragraph{Fine-grained controllable talking head synthesis.} An example of our fine-grained control over synthesized talking head is in~\cref{fig:teaser}. For a given appearance reference, we can control its lip motion, head pose, eye motion, and expression, via different respective audio signals and driving frames, and composite all motions to synthesize a vivid talking head. Our method largely improves the controllability of talking head synthesis upon previous methods, where they cannot achieve separate control over all these factors. More visual results are in the supplementary materials.

\begin{figure}[t]
  \centering
  \includegraphics[width=1\linewidth]{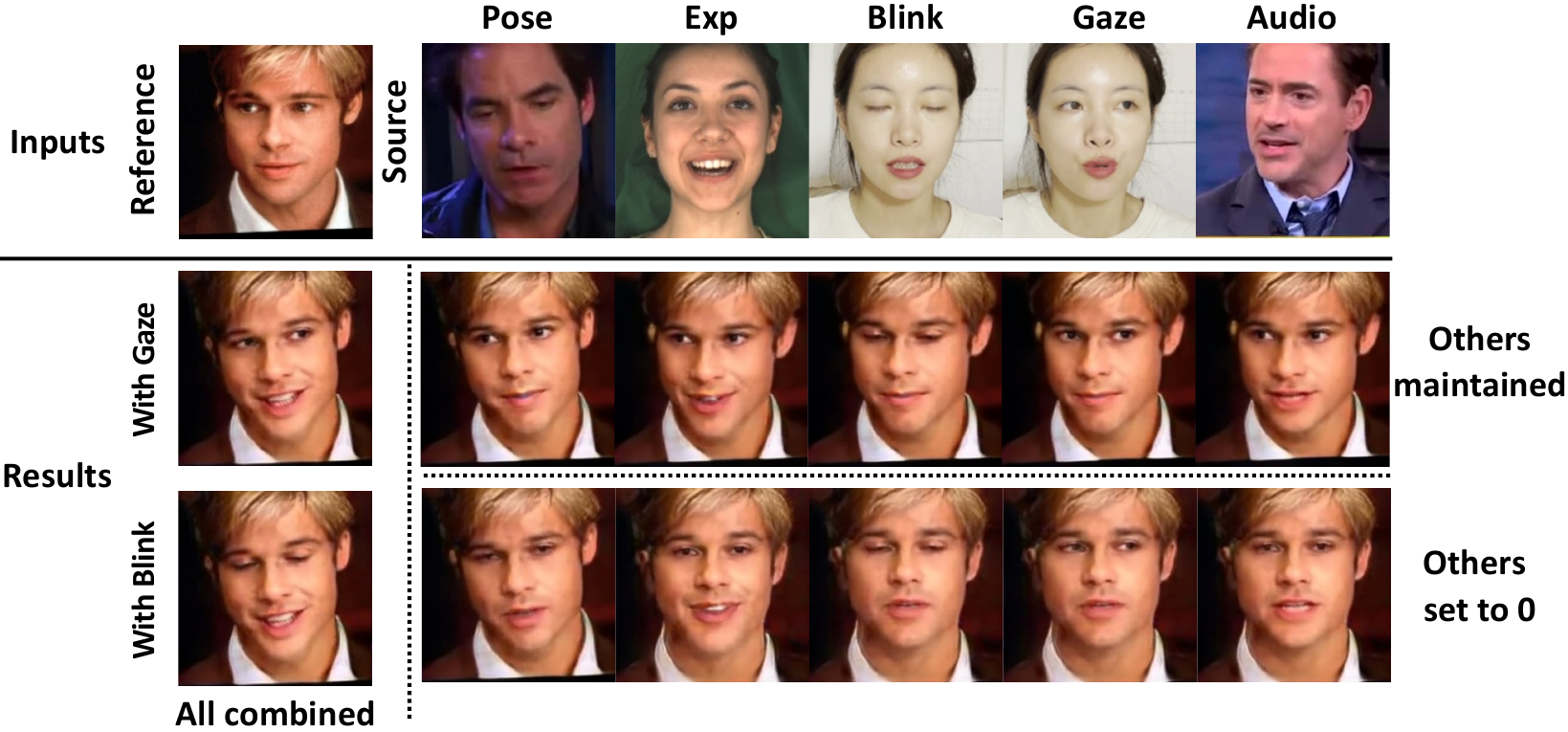}
    
   \caption{Illustration of our factor disentanglement. \textbf{Top:} Controlling one factor while leaving the others unchanged. \textbf{bottom:} Controlling one factor and setting the others to zeros.}
   \label{fig:dis_quality}
\end{figure}

\paravspace
\paragraph{Disentangled controllability.}
We demonstrate factor disentanglement of our method by changing one motion factor at a time given different driving signals in \cref{fig:dis_quality}. Our method can independently control the motion of each property to mimic the driving source, and leave all other properties unchanged. Moreover, we can also set all motions to their canonical positions (\ie set features to zero) except the motion to be controlled. These enable our method for diverse downstream applications with different requirements.

\begin{figure}[t]
  \centering
  \includegraphics[width=0.88\linewidth]{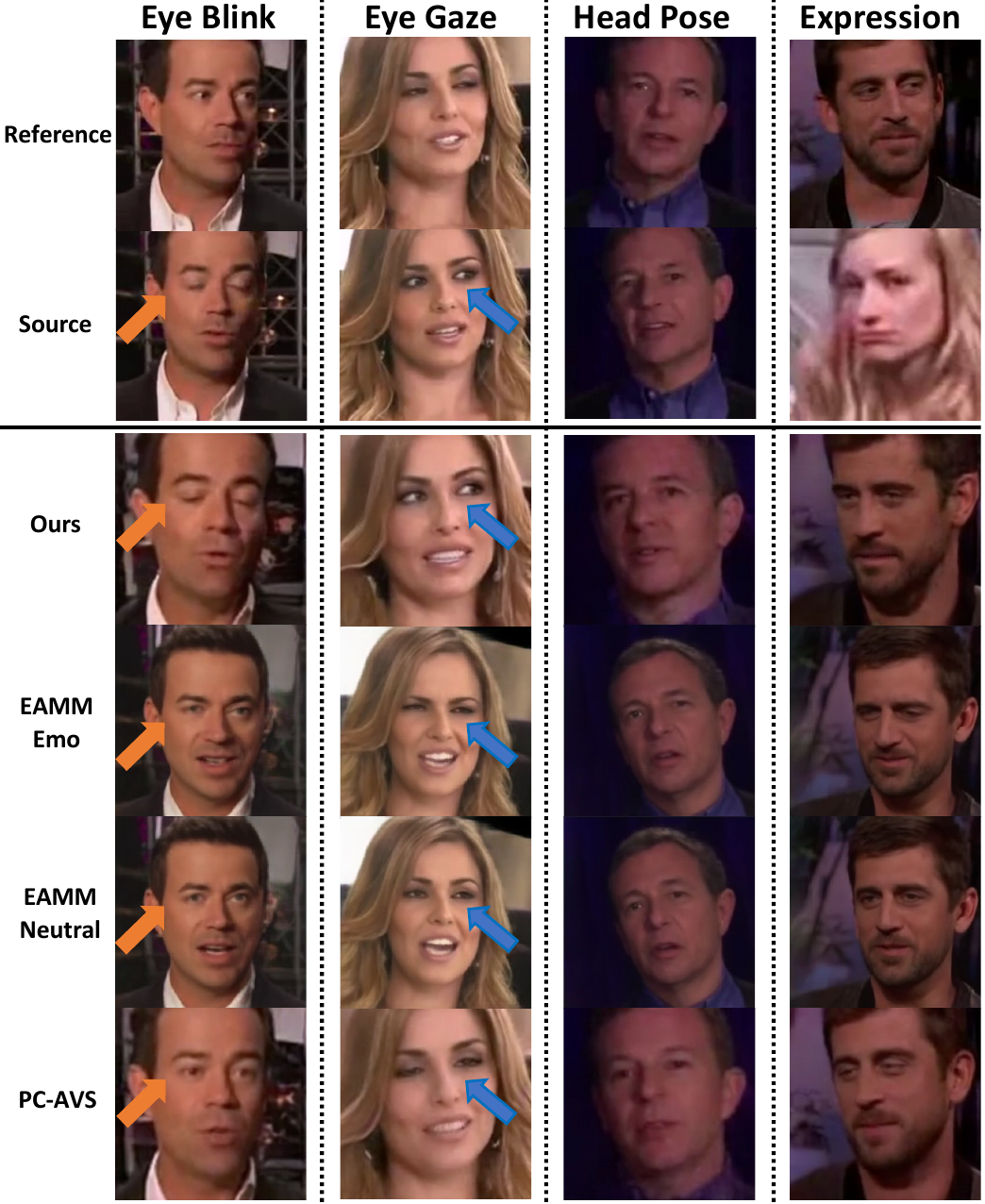}
    
   \caption{Visual comparison with other methods. The first three columns show the self-driving results. The last column shows the cross-video results. The captions above show the factors that should be focused on in each column.}
   \label{fig:nonlipmotion_compare}
\end{figure}

\paravspace
\paragraph{Comparison with prior art.}
We show visual comparisons between our method, PC-AVS~\cite{zhou2021pose} and EAMM~\cite{ji2022eamm} in \cref{fig:nonlipmotion_compare}. For eye motion and head pose, we adopt the same self-driving setting as in \cref{sec:quant}. For expression, we use the cross-video setting. We leave the lip motion comparison in the supplementary materials. As shown, our method can well mimic different motions in the driving frames compared to the other methods.
PC-AVS~\cite{zhou2021pose} can only control head pose besides lip motion, so their synthesized faces have different eye motions or expressions compared to the driving frames. It also shows inferior results in head pose control as depicted by the second right column in \cref{fig:nonlipmotion_compare}, due to using an implicit head pose representation instead of the explicit 3D rotation and translation. 
EAMM~\cite{ji2022eamm} cannot well control head pose and eye motions. Moreover, although it can control the expression of a synthesized face, its produced expression is different from the driving source as shown in the last column in \cref{fig:nonlipmotion_compare}.

\subsection{User Study}
We further conduct user studies for a more comprehensive evaluation. We ask the participants to score from 1 to 5 for the quality of different properties in the synthesized images (5 is the best).
The results are in \cref{table_user_study}. Our method achieves the best result in expression control quality and facial motion naturalness. And we achieve the second-best result on lip motion synchronization and get very close to the best one (\ie PC-AVS). More details and a user study of factor disentanglement are in the supplementary materials.

\subsection{Ablation Study}
We conduct an ablation study to validate the efficacy of our proposed feature-level decorrelation in the expression disentanglement stage. We conduct a similar self-driving experiment as in~\cref{sec:quant}, except that we use the first frame in a driving video clip as the expression driving signal instead of using the expression of each frame. The corresponding frames and the audio in the same driving video clip are still used as ground truth to calculate the metrics. In theory, if the expression is well disentangled with other motions, fixing the expression source instead of using the expression in each frame will not influence the lip motion accuracy and should maintain low NLSE-C and LMD$_{m}$. As shown in \cref{table_ablation_dis}, introducing the two decorrelation strategies significantly lowers the quantitative metrics, which indicates better factor disentanglement. And leveraging both of them leads to the best result. We further conduct the cross-video driving experiment similar to~\cref{sec:quant} to evaluate the expression control accuracy of different alternatives. Our final solution only slightly decreases the expression control accuracy but leads to a large improvement in expression and lip motion disentanglement.

We also study the influence of the window size of the in-window decorrelation in~\cref{table_ablation_winsize}. A window size of 13 yields the best result which is used as our final solution.

\begin{table}[]\small
\centering
\caption{Ablation study on expression disentanglement.}
\setlength{\tabcolsep}{1mm}{
\begin{tabular}{@{}c|ccc|cc@{}}
\toprule
\multicolumn{1}{c}{\multirow{2}{*}{Method}} & \multicolumn{3}{c}{Voxceleb2} & \multicolumn{1}{c}{Voxceleb2} & \multicolumn{1}{c}{Mead}     \\
\multicolumn{1}{c}{} & LSE-C$\uparrow$ & NLSE-C$\downarrow$ & LMD$_m\downarrow$ & Exp$\downarrow$ & Exp$\downarrow$ \\
\midrule
  
  No dis &   3.78  &   0.486  & 1.81 & \bf{0.151} & 0.178 \\
  \hline
  + In-win &   \bf{7.60}  &  0.034  &  1.76 & 0.159   & 0.178 \\
  + Decorr &  7.02   &   0.045    & 1.66 & 0.157  & \bf{0.173} \\
  All  &   7.30  &  \bf{0.007} &  \bf{1.27}   &  0.163 & 0.179 \\
      
\bottomrule

\end{tabular}}
\label{table_ablation_dis}
\end{table}

\begin{table}[]\small
\centering
\caption{Ablation study on window size of the in-window decorrelation strategy. 
}
\setlength{\tabcolsep}{1mm}{
\begin{tabular}{@{}c|ccc|cc@{}}
\toprule
\multicolumn{1}{c}{\multirow{2}{*}{Size}} & \multicolumn{3}{c}{Voxceleb2} & \multicolumn{1}{c}{Voxceleb2} & \multicolumn{1}{c}{Mead}     \\
\multicolumn{1}{c}{} & LSE-C$\uparrow$ & NLSE-C$\downarrow$ & LMD$_m\downarrow$ & Exp$\downarrow$ & Exp$\downarrow$ \\
\hline
  
  7 &   7.01  &   0.046  & 1.81 & \bf{0.163} & 0.179 \\
  13 &   \bf{7.30}  &  \bf{0.007}  &  \bf{1.27} & \bf{0.163}   & 0.179 \\
  25 &  7.23   &   0.016    & 1.32 & 0.164  & \bf{0.178} \\
      
\bottomrule

\end{tabular}}
\label{table_ablation_winsize}
\end{table}

\section{Conclusion}
We presented a fine-grained controllable talking head synthesis method. The core idea is to represent different facial motions via disentangled latent representations. A progressive disentangled representation learning strategy is introduced to separate individual motion factors in a coarse-to-fine manner, by exploring the inherent properties of each factor in unstructured video data. Experiments demonstrated the efficacy of our method on disentangled and fine-grained control of diverse facial motions.

\paravspace\paragraph{Limitations.} Our method mainly focuses on disentangled motion control. The synthesized images may lack fine details and we leave their improvement as future works.

{\small
\bibliographystyle{ieee_fullname}
\bibliography{egbib}
}

\clearpage
\appendix

\begin{strip}
\centering
\Large{\textbf{Supplementary Material}}
\end{strip}

\renewcommand{\thesection}{\Alph{section}}
\renewcommand{\thefigure}{\Roman{figure}}
\renewcommand{\thetable}{\Roman{table}}
\renewcommand{\theequation}{\Roman{equation}}
\setcounter{figure}{0}
\setcounter{equation}{0}

\section{More Implementation Details}
\label{sec:implementation}

\subsection{Data Preparation} \label{sec:data}
We train our method on all available videos in the training split of VoxCeleb2~\cite{Chung18b} dataset. For evaluation, we use the test split of both VoxCeleb2 and Mead~\cite{wang2020mead} dataset. We randomly sample $500$ test video clips from VoxCeleb2, and $460$ test clips from the Mead following the official setting.

All video frames are aligned following the official annotations~\cite{Chung18b}, and then resized and center-cropped to $224\times224$.
Corresponding audios are extracted from the original videos by ffmpeg,  and then processed with a sample rate of $16,000$ and converted to Mel-spectrograms via FFT. The window size, hop size, and the number of Mel bands are set to $1,280$, $160$ and $80$, respectively.

\subsection{More Training Details}
\label{sec:training_details}

\paragraph{Appearance and motion disentanglement.} 
We follow~\cite{burkov2020neural} to learn the appearance encoder $E_{app}$, motion encoder $E_{mot}$, and the extra image generator $G_0$. Different from \cite{burkov2020neural}, for the appearance encoder, we send a single appearance reference as input to obtain the appearance latent feature during training, instead of taking the average latent feature of multiple appearance frames in a video clip. Apart from the original training losses proposed in~\cite{burkov2020neural}, we further introduce a motion reconstruction loss as described in Sec.~3.1 in the main paper (\ie Eq.~(1)). We set the initial learning rates for $E_{app}$, $E_{mot}$ to $5e^{-5}$. The initial learning rates for $G_0$ and an extra discriminator for computing the adversarial loss in~\cite{burkov2020neural} are set to $5e^{-5}$ and $5e^{-6}$, respectively. The learning rates of all networks are decayed by a rate of $0.5$ for every $80,000$ iterations. We trained the whole pipeline with a batchsize of $24$ for $50$ epochs on $8$ Tesla V100 GPUs with $32$GB memory, which took around $2$ weeks.

\paragraph{Lip motion disentanglement.} We adopt the audio-visual contrastive learning scheme~\cite{zhou2021pose} to learn the lip motion encoder $E_{lip}$ and the audio encoder $E_{aud}$. The two models are trained on audio-video pairs with the InfoNCE loss~\cite{oord2018representation} as described in the main paper (\ie Eq.~(2) and (3)). The original training scheme in~\cite{zhou2021pose} utilizes frames from the videos different from those deriving the audio signals to construct the negative pairs, which we found can learn non-lip motion information in the obtained lip motion features. Therefore, we only use the unsynchronized frames and audio from the same video clip as the negative pairs during training. We set the initial learning rates of $E_{lip}$ and $E_{aud}$ to $1e^{-5}$, with a decay rate of $0.93$ by every $200,000$ iterations. We train the two networks with a batchsize of $32$ for $30$ epochs. Each item in a batch contains $1$ positive pairs and $8$ negative pairs. The training took $2$ days on $4$ Tesla V100 GPUs.

\begin{figure}[t]
  \centering
  \includegraphics[width=0.8\linewidth]{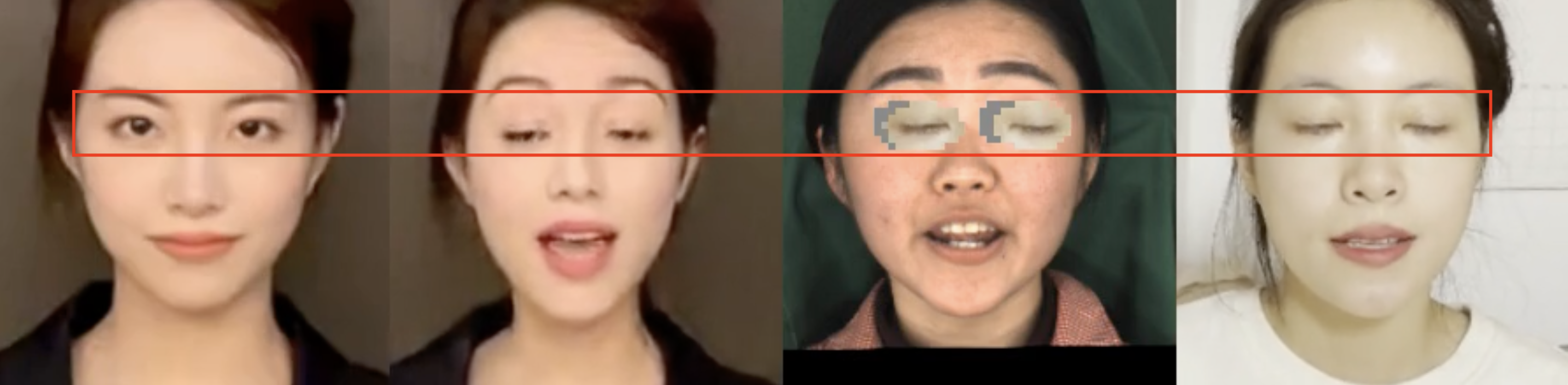}
  \includegraphics[width=0.8\linewidth]{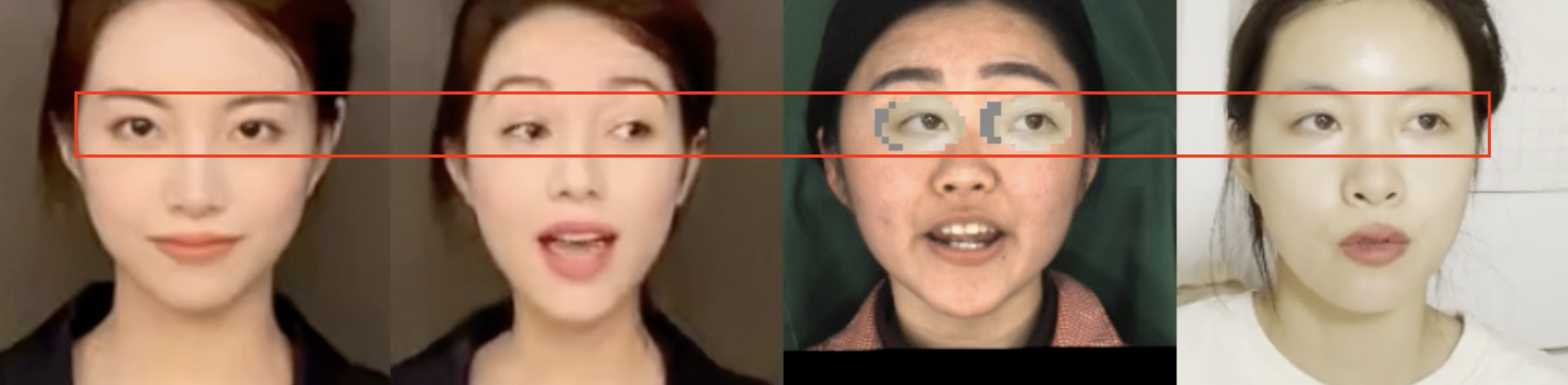}
    
  \caption{Our observation on disentangled eye motion control in the face-reenactment setting in our appearance and motion disentanglement stage. The first column is the appearance reference, the second column is the reenactment result, the third column is the driving source where the eye region comes from the images in the last column. As shown in the figure, the eye motion can be controlled independently without affecting the lip motion in this scenario, which inspires us to design the eye-motion contrastive learning.}
  \label{fig:eye_observe}
  \vspace{-5pt}
\end{figure}

\paragraph{Eye motion disentanglement.} The eye motion encoder $E_{eye}$ is learned using our proposed eye-motion contrastive learning described in Sec.~3.2 in the main paper. We describe more details about the motivation behind. 
Specifically, since our first stage is based on the face reenactment method of~\cite{burkov2020neural}, we can already synthesize a talking face with the unified motion feature of a driving frame and a given appearance feature via the image generator $G_{0}$. We find that by simply replacing the eye region of the driving frame with a new one bearing different eye blink and gaze, we can achieve a disentangled control of eyes in the synthesized face without affecting other facial motions, as shown in~\cref{fig:eye_observe}. 
Inspired by this, we formulate the eye-motion contrastive loss in the main paper.

We set the initial learning rate for $E_{eye}$ to $1e^{-5}$, decayed by a rate of $0.5$ for every $80,000$ iterations. The network is trained with a batchsize of $128$ for $30$ epochs. The training took $2$ days on $4$ Tesla V100 GPUs.

\paragraph{Head pose disentanglement.} The head pose encoder $E_{pose}$ is learned by regressing the pseudo pose labels as depicted in Sec.~3.2 in the main paper. The learning rate of $E_{pose}$ is also set to $1e^{-5}$ with a decay rate of $0.5$ by every $80,000$ iterations. The network is trained with a batchsize of $128$ for $30$ epochs similar to the eye motion encoder. The training took $2$ days on $4$ Tesla V100 GPUs.

\paragraph{Expression disentanglement.} The expression encoder $E_{exp}$ and our final image generator $G$ are learned via our proposed feature-level decorrelation and complementary self-reconstruction in Sec.~3.3 in the main paper. During this stage, all other networks are fixed, including $E_{app}$, $E_{mot}$, $E_{lip}$, $E_{aud}$, $E_{eye}$, and $E_{pose}$.
For the in-window decorrelation, we set the window size to $13$. For the lip-motion decorrelation, we set the memory bank size to $512$ for an accurate estimation of the feature correlation. 
We set the initial learning rates to $1e^{-5}$ and $2e^{-5}$ for $E_{exp}$ and $G$, respectively. The learning rate for an extra discriminator to compute the adversarial loss is set to $3.5e^{-6}$. 
The expression encoder is trained during the first $40,000$ iterations and frozen for the following steps. The learning rates for the generator and the discriminator are decayed with a rate of $0.5$ by every $80,000$ iterations. We use a batchsize of $16$ and train all networks for $50$ epochs. It took $2$ weeks on $8$ Tesla V100 GPUs.

\subsection{Quantitative Evaluation Details}
In Sec.~4.1 in the main paper, we conducted multiple experiments for quantitative metrics calculation (\ie Tab.~1 and 2 in the main paper) under two different settings, namely the \textit{self-driving setting} and the \textit{cross-video setting}.

In the self-driving setting, we use all test clips described in \cref{sec:data} for evaluation. We set the first frame in each video as the appearance reference, and drive it using the video frames and the corresponding audio from the same video clip. The audio signals are used to drive the lip motion and the video frames for other motions. Since the source and the target are from the same video, we can easily use the driving frames as the ground truth to evaluate the performance of each method.

In the cross-video setting, we use the first frame from $100$ randomly sampled test video clips as an appearance reference and use the first frame from another $100$ random test video clip as the driving frame to control all non-lip motions. We still use the audio signals from the video clip of the corresponding appearance frame to control the lip motion.
The cross-video setting is designed to evaluate the expression control performance, where we extract the expression parameters of the synthesized videos and their corresponding driving frames using a 3D face reconstructor~\cite{danvevcek2022emoca}, and compare the expression parameter difference. This helps us to evaluate if a method can precisely transfer the expression from a source to a target.
By contrast, in the self-driving setting, since the source and the target are from the same video clip, their expressions are usually the same. Under this circumstance, if a method well mimics the expression motion of the appearance reference, it is difficult to judge whether it successfully transfers the source expression to the target or merely copies the expression from the appearance reference.

\begin{table*}[]\small
\centering
\caption{Quantitative evaluation on factor disentanglement of our method. In each row, we compute the variance of a motion feature extracted from the synthesized videos when controlling different individual motion factors.}
\renewcommand{\arraystretch}{1.2}
\begin{tabular}{@{}l|ccccc@{}}
\toprule
\multicolumn{1}{c}{\multirow{2}{*}{Variance}} & \multicolumn{5}{c}{Control property} \\
\multicolumn{1}{c}{} & lip  &  pose &  blink & gaze & exp \\
\hline
Speech lip motion  &  {\bf 11.24}   &  5.16   &    0.83      &     0.74      &    3.76    \\

Head pose   &  0.0091   &    {\bf0.1597}  &  0.0041 &     0.0045      &     0.0088      \\
Eye blink  &   0.00038 &   0.00389   &    {\bf 0.06657}      &     0.00089      &     0.00225   \\
Eye gaze    &  0.089   &  0.100   &     0.095     &    {\bf 0.105}      &  0.088      \\
Expression  &  3.07   &   3.07   &     2.98     &     2.93      &   {\bf 3.59 }      \\

\bottomrule
\end{tabular}
\label{table_dientangle}
\end{table*}

\subsection{User Study Details}
\label{user_study}
We conduct two user studies to evaluate the motion control performance. In the first experiment, we ask participants to evaluate the accuracy of lip motion synchronization and expression control, as well as the naturalness of all facial motions. We generate $120$ videos using $12$ random appearance references and $10$ random driving clips and randomly select $35$ synthesized videos out of them for evaluation. Fifteen participants are asked to score from 1 to 5 for the quality of different properties in the synthesized videos (5 is the best). The corresponding results are in Tab.~3 in the main paper. 

In the second experiment, we ask the same group of participants to evaluate the disentanglement controllability of our method. We generate $5$ videos using an appearance reference and $3$ randomly selected driving videos for the head pose, expression, and eye motion, respectively. In each synthesized video, only one motion factor is controlled by the driving source and all other factors remain unchanged. The participants are asked to score from 1 to 5 for the variation level of each motion in the synthesized videos (5 indicates the largest variation, and 1 means nearly unchanged). The corresponding results are in \cref{tab:user_study2} and discussed in \cref{sec:disentangle}.

\begin{table*}[t]\small
\centering
\caption{User study on factor disentanglement of our method.\label{tab:user_study2}}
\begin{tabular}{@{}c|ccccc@{}}
\toprule
\multicolumn{1}{c}{\multirow{2}{*}{\makecell[c]{Variance}}} & \multicolumn{5}{c}{Control property} \\
\multicolumn{1}{c}{} & lip & pose & blink & gaze & exp \\
\hline
lip  &  {\bf 4.7}   &  1.1   &     1    &      1     &    1.1    \\
pose   & 1.1  &    {\bf4.6}  &  1 &      1.2     &      1.3    \\
blink  &  1.1  &   1.1  &    {\bf4.1}      &      1.5     &   1     \\
gaze    & 1   &  1.1  &  1.4     &    {\bf4.4}      &    1    \\
exp  &   1.3  &  1.1   &   1   &     1    &   {\bf  3.7}      \\

\bottomrule
\end{tabular}
\label{table_dis}
\end{table*}

\begin{figure*}[t]
  \centering
  \includegraphics[width=1\linewidth]{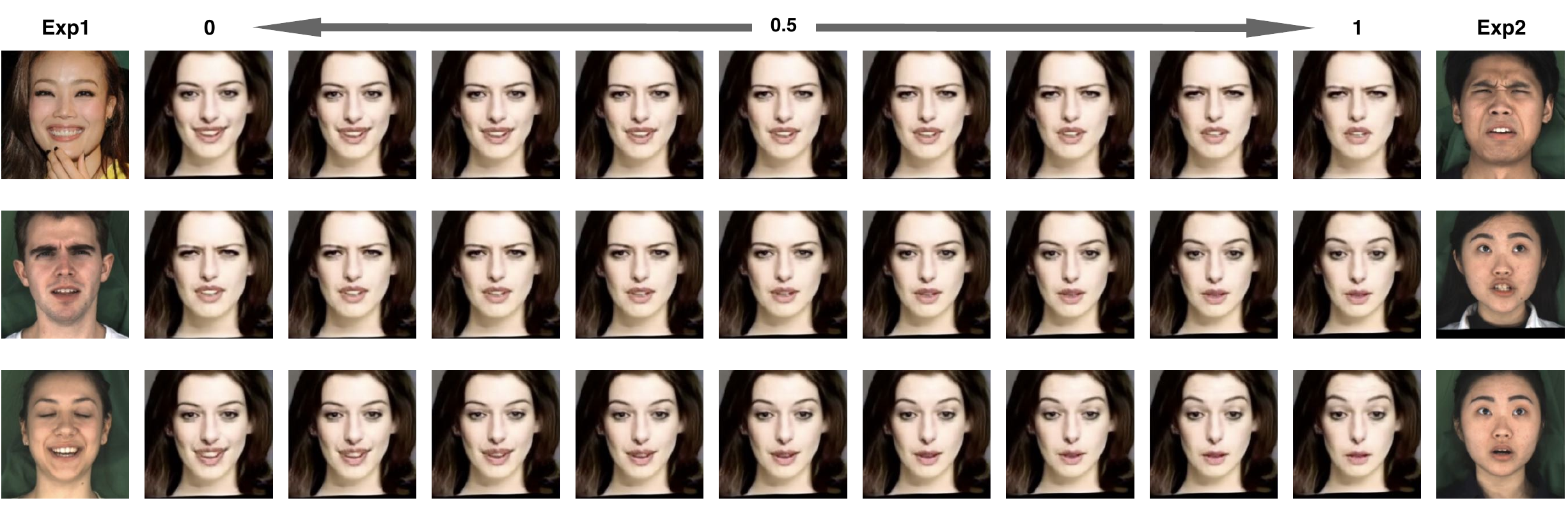}
    
   \caption{Expression interpolation by our method.}
   \label{fig:exp_interpolate}
\end{figure*}

\section{More Results}
\subsection{Fine-Grained Controllable Talking Heads}
\begin{figure*}[p]
  \centering
  \includegraphics[width=1\linewidth]{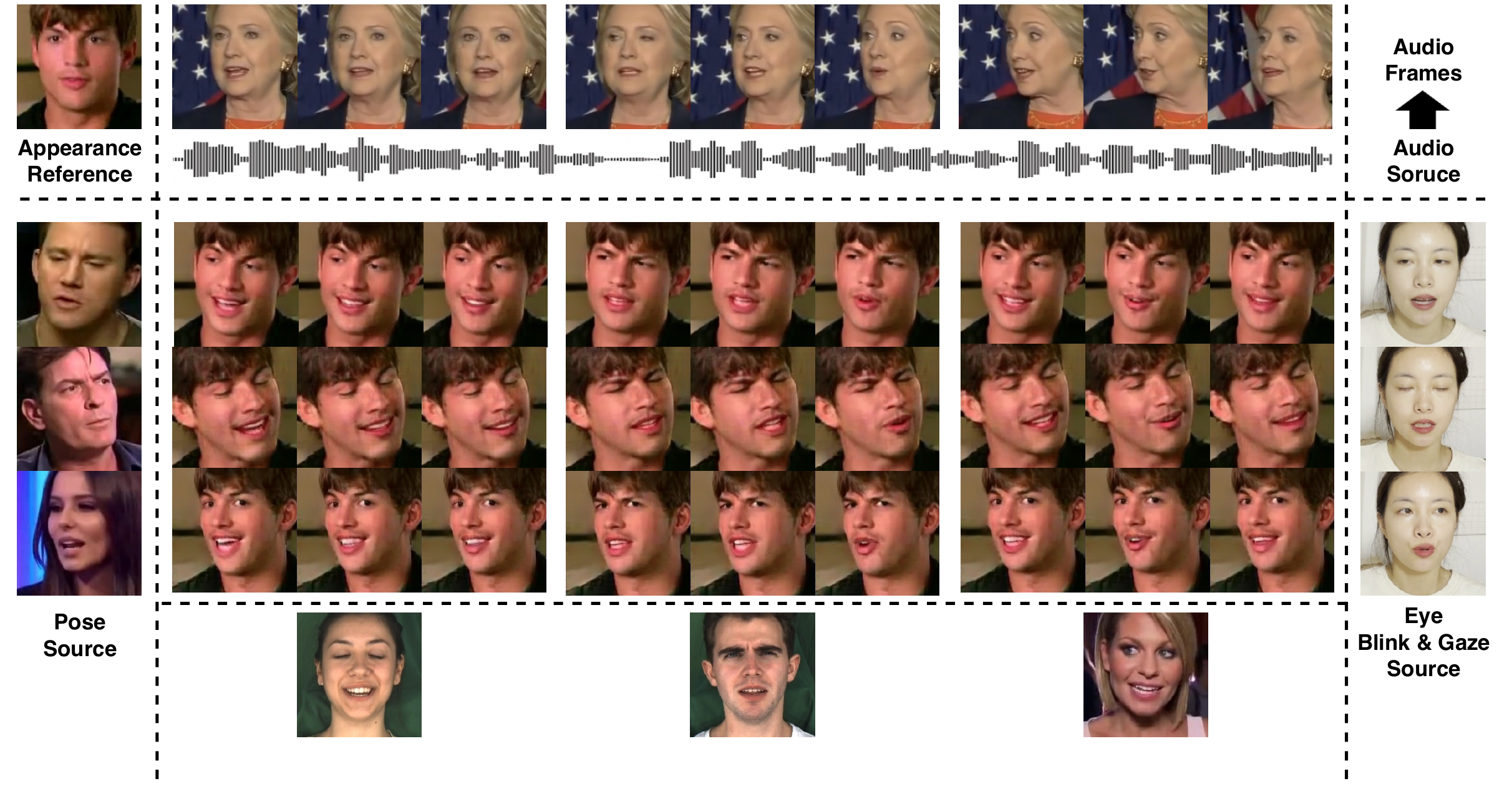}
    
   \caption{Fine-grained controllable talking heads synthesized by our method.}
   \label{fig:fine-grained_show1}
\end{figure*}

\begin{figure*}[p]
  \centering
  \includegraphics[width=1\linewidth]{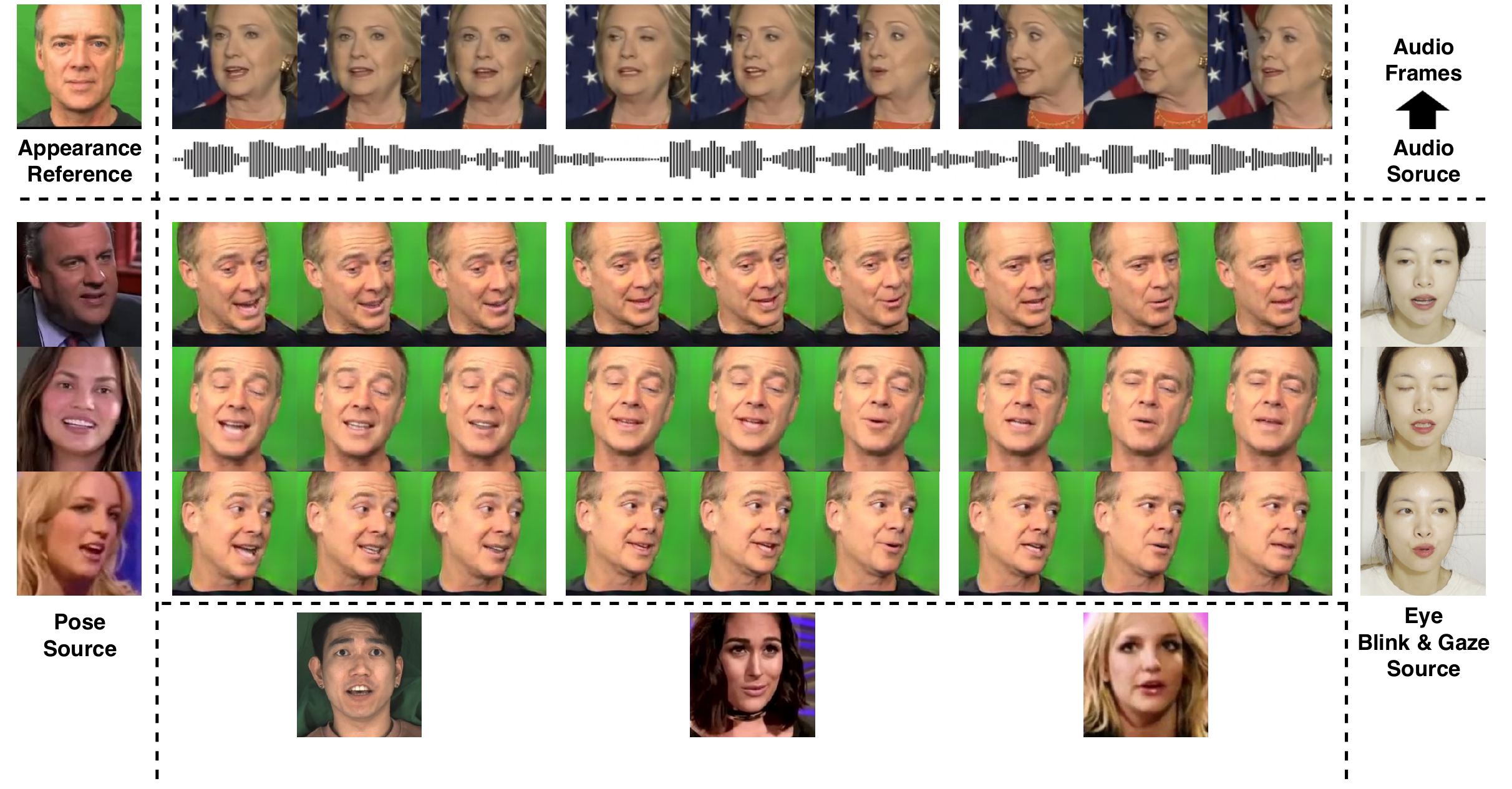}
    
   \caption{Fine-grained controllable talking heads synthesized by our method.}
   \label{fig:fine-grained_show2}
\end{figure*}

Figure~\ref{fig:fine-grained_show1} and~\ref{fig:fine-grained_show2} show more talking head synthesis results by our method. Our method well mimics the motions from different driving sources and combines them to generate vivid talking heads. \textbf{Animations can be found in the accompanying video.}

\subsection{Disentangled Controllability} \label{sec:disentangle}
We quantitatively evaluate the disentangled controllability of our method. To this end, we generate talking head images by only varying one motion factor and setting other factors to zeros (\ie canonical positions). We then extract corresponding motion features from the synthesized results and compute the variance of each motion factor in a video clip. Ideally, if different motions are perfectly disentangled, the computed variances will be close to zero for all motions except the one being controlled. 

In practice, we use off-the-shelf models to extract each motion feature from our synthesized images. For eye gaze and blink, we use the model of~\cite{fischer2018rt}. For expression and pose, we use a 3D face reconstruction model~\cite{deng2019accurate}. For lip motion, we use the model of~\cite{chung2016out}. The variance of each motion factor $\bm{f}$ is computed using the following equation:
\begin{equation}
  var(\bm{f}) = \frac{1}{N}\sum_{i=1}^N{\frac{1}{M_i}\sum_{j=1}^{M_i}{\lVert\bm{f}_{ij} - \bm{\Bar{f}}_{i}\rVert_2}},
  \label{eq:variance}
\end{equation}
where $\bm{f}_{ij}$ is the corresponding extracted motion feature of the $j$-th frame in the $i$-th video clip, $\bm{\Bar{f}}_{i}$ is mean of $\bm{f}_{ij}$, $N$ is the number of test videos, and $M_i$ is the length of each video clip.

Table~\ref{table_dientangle} shows the computed variance of each motion factor. Each row shows the variance of a single motion factor under different motion control. As shown, the variance of a factor reaches the maximum when the controlling factor is the same with it, and largely decreases when controlled under a different motion factor. This indicates that our method can disentangle different motion controls so that they have a minor influence on each other. 

However, the computed variance can still be large in some cases (\eg the left four columns in the last row in \cref{table_dientangle}). This is due to that the off-the-shelf motion feature extractors are not perfect and can be influenced by variations of other motions when extracting a certain motion feature. Therefore, we refer the readers to the accompanying video to examine the disentanglement ability of our method. We also conduct a user study to better evaluate the factor disentanglement. The results are in \cref{tab:user_study2} (see \cref{user_study} for detailed description). As shown, the variance score is close to $5$ when the factor for variance calculation and the factor to be controlled are the same, and close to $1$ when they are different, which reveals the disentangled controllability of our method.

\subsection{Expression Interpolation}
We further investigate the expressive ability of our learned expression feature. We show expression interpolation results by linearly interpolating two expression features from different expression sources. As shown in~\cref{fig:exp_interpolate}, our method can smoothly transfer between two different expressions. The synthesized images at interpolated points also have natural expressions. This indicates that our method learns a reasonable expression latent space that supports continuous expression control. 

\begin{figure}[t]
  \centering
  \includegraphics[width=1\linewidth]{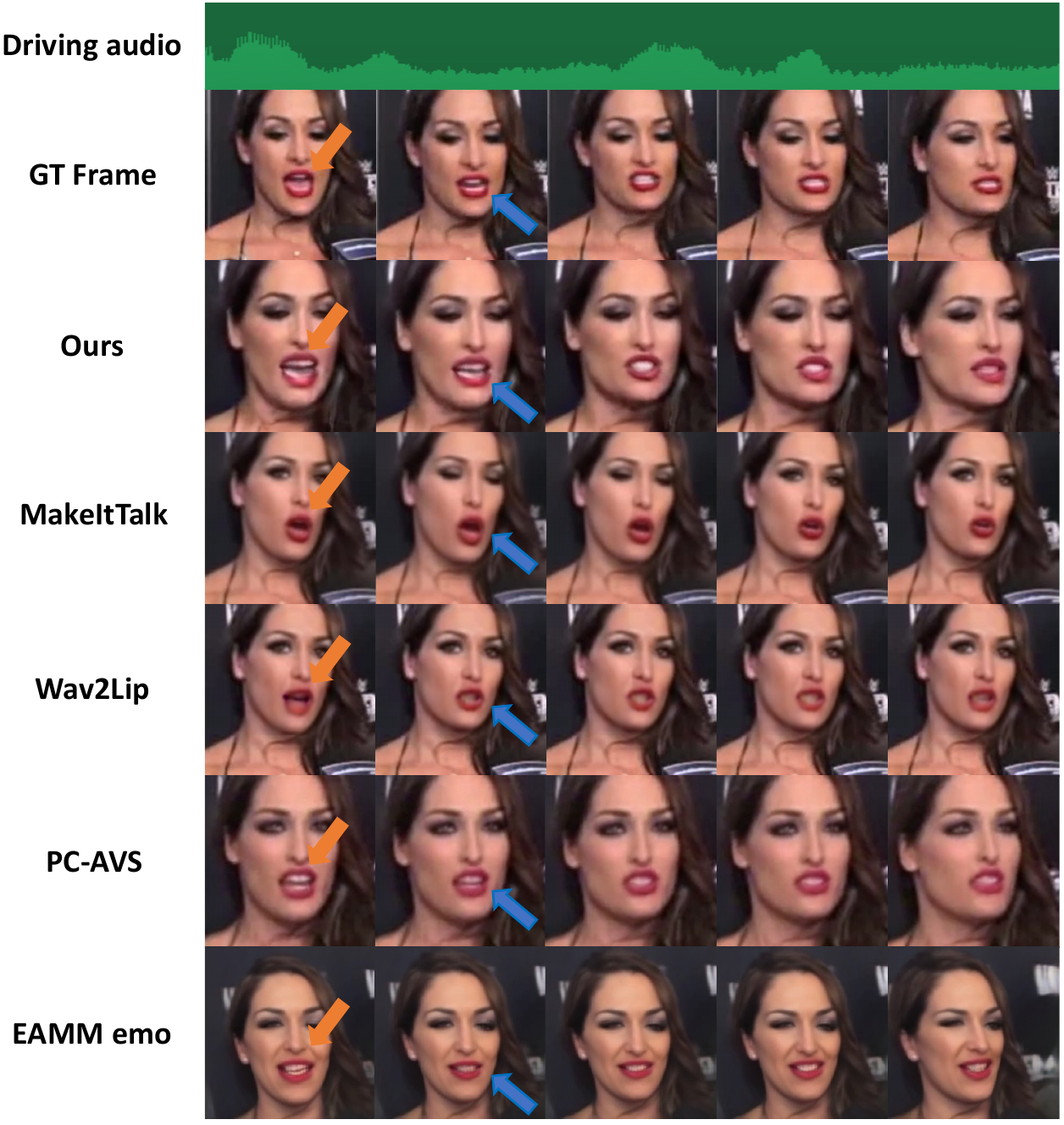}
    
   \caption{Comparison on lip motion control. Images are synthesized under the self-driving setting where the lip motion is driven by the audio signal.
   Our method yields the best result.}
   \label{fig:lipmotion_compare}
\end{figure}

\subsection{Comparison with the prior methods}
We show the lip motion synthesis comparison in~\cref{fig:lipmotion_compare}. The images are synthesized under the self-driving setting. As depicted, the lip motion generated by our method is natural and closer to the ground truth compare to the alternatives. \textbf{See the accompanying video for animations.}

\subsection{Ablation Study}
\paragraph{Motion reconstruction loss.}

\begin{figure*}[p]
  \centering
  \includegraphics[width=1\linewidth]{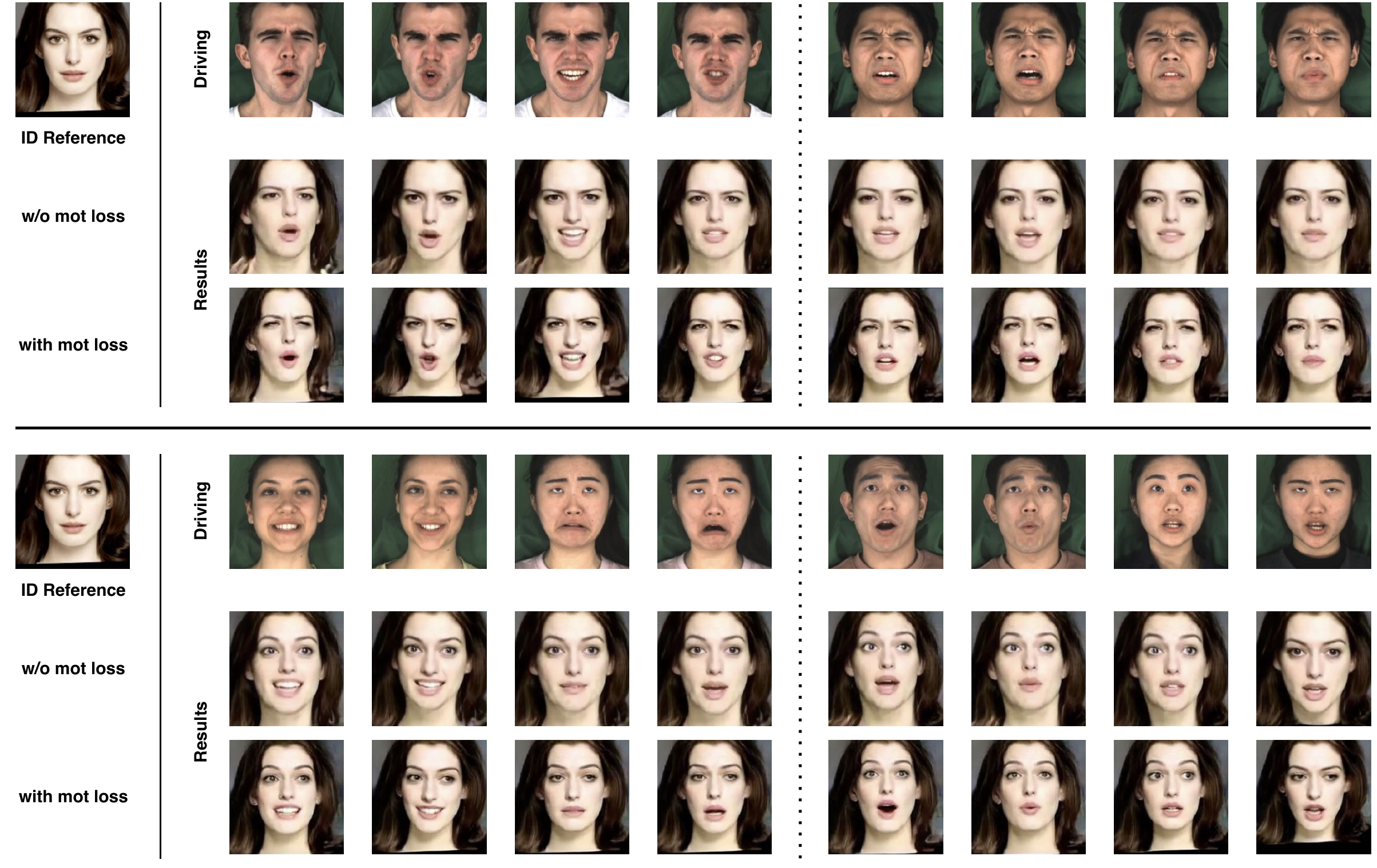}
    
   \caption{Ablation study on the motion reconstruction loss in the appearance\&motion disentanglement learning.}
   \label{fig:mot_com}
\end{figure*}

We further conduct an ablation study to validate the efficacy of our motion reconstruction loss proposed in Sec.~3.1 in the main paper.
As shown in~\cref{fig:mot_com}, with the motion reconstruction loss, facial motions in the synthesized images contain more details and are closer to the driving sources. By contrast, removing the motion reconstruction loss leads to poor reenactment results for driving sources with rich expressions. As a result, the motion reconstruction loss is important for obtaining an informative unified motion feature to achieve accurate motion control.

\section{Ethics Consideration}
Our method enables precise and disentangled control
over multiple facial motions for vivid talking head generation. While the major goal of it is to synthesize virtual avatar for applications like live streaming, it can be misused to create deceptive and harmful content of real people. Especially, one may use it to synthesize fake videos of celebrities. We do not condone using our method for generating misleading information that could harm people's reputations. We also suggest investigating advanced forgery detection methods to identify the synthesized fake images and videos to prevent illegal usage.

\clearpage

\clearpage

\end{document}